# Generative AI Is Not Ready for Clinical Use in Patient Education for Lower Back Pain Patients, Even With Retrieval-Augmented Generation

Yi-Fei Zhao[1,2], Allyn Bove, PhD, DPT[3], David Thompson, DPT[3,4], James Hill, DPT[3,5], Yi Xu[1], Yufan Ren[1], Andrea Hassman[3], Leming Zhou, PhD[1,6], Yanshan Wang, PhD[1,6,7,8]

[1]Department of Health Information Management, University of Pittsburgh, Pittsburgh, PA; [2]Stern School of Business, New York University, New York, NY; [3]Department of Physical Therapy, University of Pittsburgh, Pittsburgh, PA; [4]Alliance Physical Therapy, Murrysville, PA; [5]Athletico Physical Therapy, Pittsburgh, PA; [6]Intelligent Systems Program, University of Pittsburgh, Pittsburgh, PA; [7]Clinical and Translational Science Institute, University of Pittsburgh, Pittsburgh, PA; [8]University of Pittsburgh Medical Center, Pittsburgh, PA

**Abstract**

*Low back pain (LBP) is a leading cause of disability globally. Following the onset of LBP and subsequent treatment, adequate patient education is crucial for improving functionality and long-term outcomes. Despite advancements in patient education strategies, significant gaps persist in delivering personalized, evidence-based information to patients with LBP. Recent advancements in large language models (LLMs) and generative artificial intelligence (GenAI) have demonstrated the potential to enhance patient education. However, their application and efficacy in delivering educational content to patients with LBP remain underexplored and warrant further investigation. In this study, we introduce a novel approach utilizing LLMs with Retrieval-Augmented Generation (RAG) and few-shot learning to generate tailored educational materials for patients with LBP. Physical therapists manually evaluated our model responses for redundancy, accuracy, and completeness using a Likert scale. In addition, the readability of the generated education materials is assessed using the Flesch Reading Ease score. The findings demonstrate that RAG-based LLMs outperform traditional LLMs, providing more accurate, complete, and readable patient education materials with less redundancy. Having said that, our analysis reveals that the generated materials are not yet ready for use in clinical practice. This study underscores the potential of AI-driven models utilizing RAG to improve patient education for LBP; however, significant challenges remain in ensuring the clinical relevance and granularity of content generated by these models.*

## Introduction

Low back pain (LBP) is a common musculoskeletal condition that is responsible for approximately 65 million years lived with a disability (YLD) with a prevalence of 568 million people globally [1]. Effective management of LBP often involves rehabilitation care, including exercise, manual therapy, and/or patient education. Patient education helps patients understand their condition and learn self-management techniques to improve their functional outcomes [2]. Traditional patient education for LBP often relies on standardized materials and generalized advice, which may inadequately address the needs and concerns of all patients. A large proportion of education materials are also written at reading levels exceedingly higher than the recommended levels [3], limiting their effectiveness. Artificial intelligence (AI) may offer promising results in providing more personalized education materials. By leveraging AI, healthcare providers can better tailor educational content to individual patients and potentially improve understanding and self-management.

AI technologies have improved patient education across various conditions by providing more accessible, personalized, and interactive educational resources through chatbots and virtual health assistants. These AI-powered tools can deliver information in a conversational format and analyze patient data to identify patterns and specific needs of patients in dentistry [4] or diabetes management [5]. While these approaches have shown promise, they are limited to their reliance on predefined rules and static guidelines. They are unable to adapt to the nuances of evolving individual LBP conditions, lacking depth and variety of information to address the complexities of LBP. Furthermore, AI-generated patient education has yet to be studied in the context of chronic LBP, leaving a gap in the exploration of its potential for this condition.

Recent advancements in large language models (LLMs) and generative artificial intelligence (GenAI) offer a promising approach to overcoming the limitations of traditional AI in patient education. LLMs such as OpenAI's GPT series [6], Google's BERT [7], and Meta's Llama models [8] have demonstrated remarkable capabilities in understanding and generating human-like text. These models can provide more nuanced and context-aware patient education materials by leveraging large amounts of data and sophisticated language processing techniques. Recent efforts in other medical conditions, such as urolithiasis [9] and uveitis [10], use LLMs and GenAI to create patient education materials or rewrite them at improved readability levels. However, these models are not fine-tuned for specific medical conditions, resulting in inaccuracies or incomprehensiveness in generated materials. Our study addresses these gaps by leveraging Retrieval-Augmented Generation (RAG) [11] to combine both the strengths of LLMs/GenAI with targeted information retrieval. RAG enhances traditional LLM capabilities by retrieving relevant documents from a custom external knowledge base and using them to guide generation. This ensures that the model generates contextually accurate and specific information. Knowledge-intensive tasks like patient education material generation benefit greatly from RAG [11]. LLMs can refer to an external knowledge source, ensuring the generated content is accurate, relevant, and individualized to patients with LBP [11].

This paper presents a novel approach to developing patient education materials for LBP by utilizing RAG and GenAI. Our study is the first to integrate these AI techniques specifically for LBP, as no prior studies have focused on using AI for LBP patient education. The contributions of this study are threefold: 1) demonstrating the feasibility and effectiveness of RAG in producing accurate and personalized patient education materials to be used in clinical settings; 2) highlighting the potential of LLMs to overcome the limitations of traditional AI models in the healthcare domain; and 3) providing a framework and offering insights into future applications of GenAI and RAG to improve patient education. Through this, we aim to evaluate the impact of RAG integrated with LLMs in generating patient education materials tailored to patients with LBP.

**Related Works**

The application of AI in patient education has evolved significantly. Early implementation relied on traditional AI. Educational materials were personalized based on an analysis of a patient's medical history, demographic information, and treatment records [4]. Traditional AI can identify patterns throughout patients and create educational materials to address certain needs or boost engagement and retention of materials [4]. Additionally, chatbots and virtual assistants powered by AI also provide real-time personalized guidance and recommendations to patient queries. However, traditional AI is built on rigid rule-based systems; it excels in analyzing data and recognizing patterns but cannot generate new content. Systems that create personalized education materials will take strict queries of patient data. The use of traditional AI, whether for creating educational materials or chatbots, is also limited by its accuracy and reliability [4]. Outputted content requires validation and supervision by professionals [4].

The emergence of GenAI has addressed many of these limitations. Unlike traditional AI, GenAI models, such as LLMs, can generate new content by leveraging large datasets and deep learning techniques. These models can produce highly personalized educational content tailored to individual patient contexts that align with more recent medical guidelines. Recent studies utilizing large language models for generating patient education materials have shown promising results in improving readability. Kianian et al. 2023 explored the effectiveness of GPT-4.0 and Bard in generating uveitis-related materials [10]. It is estimated that the average US resident reads at an 8th-grade level, and the average patient on Medicare reads at a 5th-grade level [12]. In their study, Kianian et al. found significantly lower readability scores in ChatGPT-produced materials when adjusting prompts to specify creating materials at a "6th-grade Flesch-Kincaid Grade Level" instead of "easy to understand by an average American" [10]. Despite advancements made by GenAI, there are still concerns about generating misleading information or "hallucinating" [13] due to an inability to understand the semantics and context of language [14]. This creates safety concerns if experts do not review and validate educational materials. Additionally, the black-box nature of deep learning models raises concerns about transparency and interpretability, which are crucial in healthcare settings. Furthermore, ensuring the ethical use of these models, especially in forms of bias, remains an ongoing issue.

**Materials and Methods**

*Data Description*

A cohort of 30 synthetic profiles of patients with low back pain was generated using ChatGPT-4o. These patient profiles were generated based on a baseline questionnaire to capture detailed information regarding patients' work status, daily activity levels, exercise routines, and beliefs and attitudes regarding exercise desk posture, lifting technique, physical therapists, injections, imaging, and bed rest. The profiles were reviewed and revised for clarity and clinical applicability by a doctoral-educated physical therapist (author AB) with 15 years of experience.

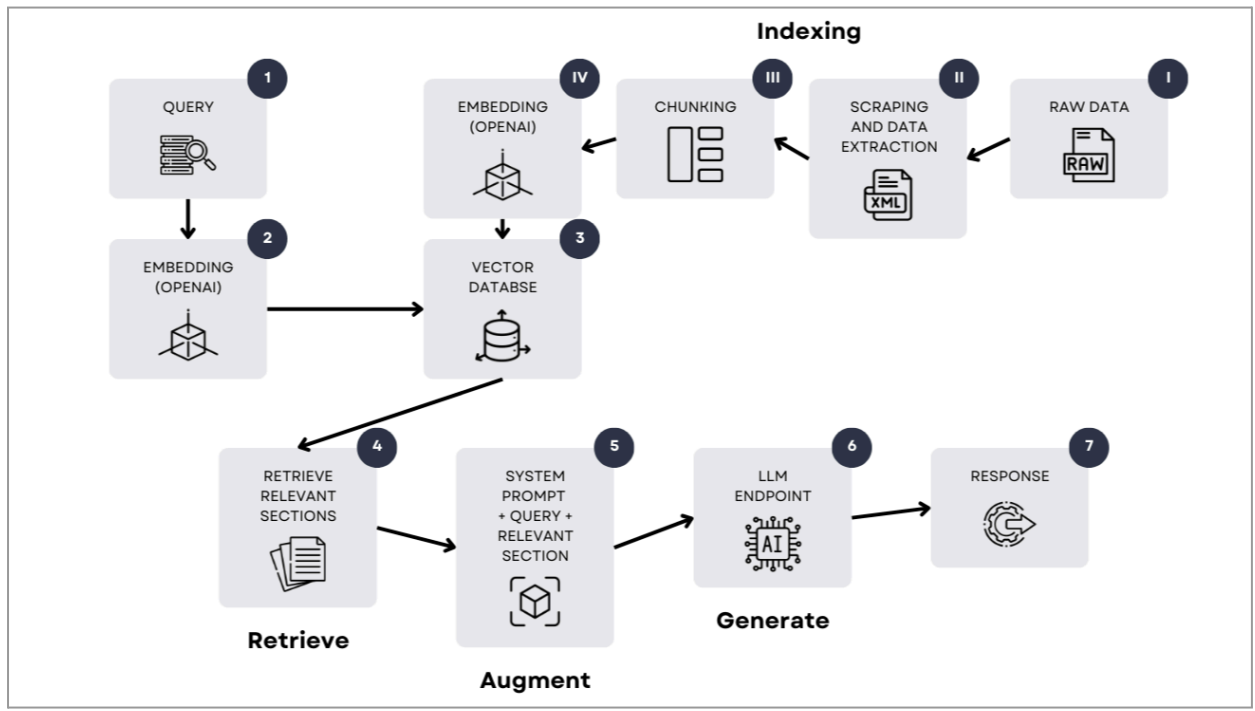

Figure 1: Overview of our RAG-based LLM approach.

*RAG Pipeline*

The core methodology of this study involves using RAG with LLMs. The RAG model combines the strengths of generative models with retrieval-based models. The pipeline (**Figure 1**) includes indexing, retrieval, augmentation, and generation. A comprehensive knowledge base is constructed during the indexing process, drawing on the physical therapist's clinical experience in delivering patient education materials. This knowledge base is a structured collection of data that is easily searchable and retrievable by LLMs to enhance the accuracy and relevance of the educational content provided to patients. Raw data (I) or unprocessed information from sources includes pages from MedlinePlus, current LBP clinical guidelines, and research articles. Relevant MedlinePlus pages and associated links on those pages are scraped and stored in XML format with sections (II). Clinical practice guidelines from various professional associations and health organizations and journal articles are also saved and sectioned in XML files (II). This knowledge base, with a total of 186 XML files with 2493 sections, gets segmented into smaller, uniformly sized chunks (III), embedded using OpenAI embeddings [15] (IV), and stored in a Chroma vector database [16] (Step 3) for retrieval purposes. During retrieval and generation, user-entered queries (Step 1) are embedded using the same OpenAI embeddings (Step 2). The retrieve compares vectors using cosine similarity, and the most relevant documents—those with vectors closest to the query vector—are retrieved from the vector database (Steps 3, 4). The top k=7 sections within the 0.40 similarity threshold are retrieved for this study. Scores nearer to 0 indicate higher relevance. Seven sections were chosen because they balance providing enough context for generating accurate and complete educational materials without overwhelming the model with too much information. A cosine similarity threshold of 0.4 ensures only the most relevant documents are retrieved. A higher threshold could dilute the quality of the response with less relevant information. The selected documents are then included with the system prompt and user query (Step 5). This augmented query is sent to the LLM endpoint (Step 6), generating a response (Step 7).

The RAG model was tested with few-shot learning [17], without few-shot learning, and without RAG nor few-shot learning to compare its impact on the quality of patient education materials produced. The LangChain [18] Python package implemented and managed the pipeline components.

*Model Selection and Test*

We selected a diverse range of five LLMs to evaluate their varied capabilities to produce patient education materials. GPT-4 and GPT-3.5-Turbo represent state-of-the-art models from OpenAI, known for their robust understanding of natural language and ability to perform well across a variety of tasks. GPT-4 is the most advanced version, designed to handle complex queries with greater accuracy and nuanced understanding, while GPT-3.5-Turbo offers a more optimized and cost-effective performance. We also included GPT-4o and GPT-4o-mini, which are optimized versions of GPT-4. These models are designed to deliver high-quality outputs with reduced computational costs, offering a balance between performance and efficiency. Additionally, we selected Llama-3-8b-Instruct, as it is an open-source model from Meta's Llama series. Including these models allows us to compare their performance across a spectrum of language understanding capabilities, response quality, and computational efficiency.

For each LLM, three configurations were tested with a prompt using the structure described in Kianian et al. 2023, explicitly asking to "please create patient education materials written at a 6th-grade level Flesch-Kincaid Grade Level". First, a RAG with Few-Shot Learning (RAGFS) configuration integrated a few-shot learning approach into the RAG model. We noticed the model with only RAG fell short of teaching patients how to lift or set up their desks properly. The few-shot prompt examples are included in **Table 1**. Next, a RAG without Few-Shot Learning

c(RAGNFS) configuration utilized only the RAG model without incorporating few-shot learning. Finally, the base LLM (NRAG) was tested without the RAG framework nor few-shot learning, relying solely on the LLMs' generative capabilities to produce patient education material.

Table 1: Few-shot prompting examples

| Example | Prompt | Output |
|---|---|---|
| 1 | Can you explain how to lift properly to avoid excessive strain on the back? | **Safe Lifting Tips:**\n1. **Get Close:** Keep the item close to your body.\n2. **Bend at the Knees:** Bend your hips and knees, not your back.\n3. **Breathe:** Don't hold your breath.\n4. **Lift with Your Legs:** Use your leg muscles.\n5. **Pivot:** Move your feet, avoid twisting your back. |
| 2 | How can a patient set up their desk ergonomically? | **Ergonomic Desk Setup Tips:**\n1. **Chair:** Support your back, knees level with hips, feet flat.\n2. **Desk:** Adequate space for legs and feet.\n3. **Monitor:** Arm's length away, eye level.\n4. **Keyboard and Mouse:** Wrists straight, hands below elbow level.\n5. **Movement:** Move around at least once per hour. |

*Scoring Criteria and Procedure*

The outputs from each configuration were manually evaluated by two experienced doctoral-educated physical therapists (authors DT and JH) using a 5-point Likert scale for redundancy, accuracy, and completeness. Redundancy measures how often the model's outputs produce repeated or unnecessary information. Lower scores indicate higher redundancy. Accuracy assesses how correctly the model outputs the intended content or answer. Higher scores indicate higher accuracy. Completeness evaluates whether the model provides enough information to be useful and informative. Higher scores indicate more comprehensive content. The reviewers were blinded to which model and configuration produced each answer. Before formal evaluation, we trained the reviewers on the use of a 5-point Likert scale to evaluate the redundancy, accuracy, and completeness of each output. Additionally, several models and configurations were excluded based on their performance. Llama3 responses (RAG with Few-Shot, RAG without Few-Shot, and Base) were excluded due to their brevity. The GPT-4o with RAG but without Few-Shot configuration was excluded because the generated content was not sufficiently detailed. The base GPT-3.5-Turbo configuration was also excluded because it provided short and generalized responses, lacking the specificity required for patient education. After excluding these configurations, our final evaluation set had 10 configurations with 30 patients each for 300 total outputs. To obtain a final score for each category, we took the scores from the two reviewers and found the mean category scores for each configuration. To compare the overall scores of the models, we added the mean scores of each category to get a total score.

An additional readability assessment was included to ensure the materials were understandable to patients of varying literacy levels. Readability was automatically assessed using Flesch-Kincaid readability tests. The Flesch Reading Ease score (FRES) determined how easily the text could be read [19]. Higher scores indicate easier material to read, and lower scores indicate harder material to read [19]. The Flesch-Kincaid Grade Level can be permuted from the Flesch Reading Ease score, which estimates the U.S. school grade level necessary to comprehend the content [19].

*Data Visualization and Statistical Analysis*

Plotly (https://plotly.com/) graphing libraries were used to create radar plots, illustrating differences between overall model scores. All statistical analyses were performed using Microsoft Excel and Minitab. Descriptive statistics were used to summarize the redundancy, accuracy, completeness, and readability scores across models. Inferential statistical tests, including ANOVA and two sample t-tests, were employed to determine significant differences between models.

**Results**

*Redundancy, Accuracy, and Completeness*

**Table 2** summarizes the performance of the 10 model configurations across three key metrics: redundancy, accuracy, and completeness. Each LLM has up to three configurations: RAGFS, RAGNFS, and NRAG. **Figure 2** visualizes redundancy, accuracy, and completeness trends for each configuration.

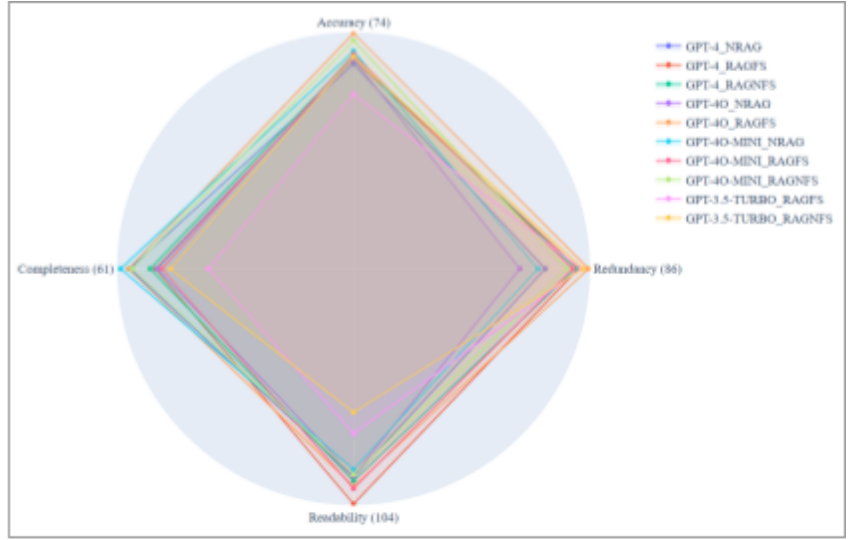

Figure 2. Comparison of different methods in terms of redundancy, accuracy, completeness, and readability.

Table 2: Performance metrics for each configuration

| Model | Redundancy (0.16) | Accuracy (0.37) | Completeness (0.87) |
| --- | --- | --- | --- |
| GPT-3.5-TURBO_RAGFS | 4.13 (1.17) | 2.73 (1.12) | 1.87 (0.68) |
| GPT-3.5-TURBO_RAGNFS | 4.17 (1.00) | 3.31 (1.00) | 2.36 (0.81) |
| GPT-4O-MINI_RAGFS | 3.98 (0.85) | 3.35 (0.73) | 2.47 (0.62) |
| GPT-4O-MINI_RAGNFS | 3.87 (0.65) | 3.57 (1.00) | 2.85 (1.15) |
| GPT-4O-MINI_NRAG | 3.35 (0.68) | 3.40 (0.98) | **3.00** (1.21) |
| GPT-4O_RAGFS | **4.27** (0.87) | **3.68** (0.89) | 2.90 (1.08) |
| GPT-4O_NRAG | 3.02 (0.60) | 3.27 (0.97) | 2.53 (0.87) |
| GPT-4_RAGFS | 4.08 (0.93) | 3.27 (0.90) | 2.55 (0.81) |
| GPT-4_RAGNFS | 4.03 (0.59) | 3.32 (1.00) | 2.62 (0.92) |
| GPT-4_NRAG | 3.48 (0.79) | 3.20 (1.10) | 2.78 (1.00) |

The highest redundancy score was observed in the GPT-4O_RAGFS model (4.27 ± 0.87), followed by GPT-3.5-TURBO_RAGNFS (4.18 ± 1.00) and GPT-3.5-TURBO_RAGFS (4.13 ± 1.17). The NRAG models, such as the GPT-4O_NRAG (3.02 ± 0.6), exhibited the lowest scores in redundancy, suggesting that models without RAG tend to produce more repetitive content. The overall standard deviation for redundancy across models was 0.16. The intraclass correlation coefficient (ICC) 95% confidence interval for redundancy was [0.29, 0.93], suggesting good reliability and agreement between evaluators [20]. ANOVA results for redundancy indicate a significant difference between models ($F(9, 587) = 14.79$, $p < 0.001$), showing that model configuration strongly influences the tendency to generate redundant information.

The GPT-4O_RAGFS model produced the highest accuracy score (3.4 ± 0.89) with GPT-4O_RAGNFS (3.57 ± 1.00) and GPT-4O-MINI_NRAG (3.40 ± 0.98) followed closely. The lowest score was associated with GPT-3.5-TURBO_RAGFS (2.73 ± 1.12). Models using few-shot learning (RAGFS) generally performed better than those without few-shot learning (RAGNFS) or the base models (NRAG). The overall standard deviation for accuracy across models was 0.37. The ICC 95% confidence interval for accuracy was [-0.57, 0.63], suggesting some reliability and agreeance between evaluators [20]. The ANOVA test for accuracy revealed statistically significant differences between the models ($F(9, 589) = 3.95$, $p < 0.001$), indicating that the use of few-shot learning and RAG impacts model accuracy.

The most complete responses were produced by the GPT-4O-MINI_NRAG model (3.00 ± 1.21), while GPT-3.5-TURBO_RAGFS (1.87 ± 0.68) showed the lowest completeness. Trends suggest that few-shot learning tends to reduce the comprehensiveness of the generated materials. The overall standard deviation for redundancy across models was 0.16. The ICC 95% confidence interval for redundancy was [-0.52, 0.68], suggesting some reliability and agreeance between evaluators [20]. ANOVA results for completeness also showed significant differences between models ($F(9, 587) = 7.34$, $p < 0.001$), confirming that model configuration affects the ability to provide complete answers.

Table 3: Readability performance averages for each configuration

| Model | FK Readability | FK Grade | Num Words | Num Syllables | Num Sentences |
|---|---|---|---|---|---|
| GPT-3.5-TURBO_RAGFS | 72.44 (11.42) | 7th Grade | 207.03 | 298.47 | 18.87 |
| GPT-3.5-TURBO_RAGNFS | 63.06 (6.24) | 9th Grade | 327.77 | 504.37 | 24.27 |
| GPT-3.5-TURBO_NRAG | 65.00 (5.60) | 9th Grade | 327.90 | 492.47 | 22.60 |
| GPT-4O-MINI_RAGFS | 96.59 (3.62) | **5th Grade** | 301.17 | 365.13 | 38.73 |
| GPT-4O-MINI_RAGNFS | 90.48 (5.11) | **5th Grade** | 409.93 | 514.90 | 40.63 |
| GPT-4O-MINI_NRAG | 88.13 (3.94) | 6th Grade | 509.70 | 653.00 | 49.63 |
| GPT-4O_RAGFS | 95.73 (4.55) | **5th Grade** | 367.77 | 444.23 | 41.53 |
| GPT-4O_RAGNFS | 87.69 (4.03) | 6th Grade | 423.27 | 539.67 | 38.40 |
| GPT-4O_NRAG | 90.90 (7.30) | **5th Grade** | 438.10 | 546.80 | 43.87 |
| GPT-4_RAGFS | **103.33** (3.560) | **5th Grade** | 306.50 | 349.40 | 44.30 |
| GPT-4_RAGNFS | 93.18 (4.03) | **5th Grade** | 321.60 | 391.80 | 30.90 |
| GPT-4_NRAG | 91.42 (5.21) | **5th Grade** | 396.47 | 492.97 | 39.53 |

*Readability*

**Table 3** outlines the readability performance of the models using Flesch-Kincaid (FK) scores, grade levels, and other text complexity measures. Models utilizing RAG, particularly RAGFS, consistently produced content at a 5th-grade reading level, with GPT-4_RAGFS producing the highest FK readability score (103.33 ± 3.56). GPT-3.5-TURBO_RAGNFS showed lower readability (63.06 ± 6.24), corresponding to a 7th-grade level. When comparing RAGFS and NRAG models on the same LLM for readability scores, each LLM showed significant differences ($p < 0.001$), supporting the conclusion that RAGFS models consistently improve readability over NRAG models. In terms of word count, the GPT-4O-MINI_NRAG model generated the longest responses (509.70 words, 653.00 syllables), while GPT-35-TURBO_RAGFS produced the shortest responses (207.03 words, 298.47 syllables).

**Discussion**

Evaluation of generated patient education materials for lower back pain revealed challenges in their clinical applicability. One key observation was that models would produce low-redundancy responses, which, while highly rated, often were generic guideline-based recommendations. When patient background information aligned with clinical practice guidelines, the model produced seemingly accurate information but offered superficial suggestions like "It's important to stay active and continue your usual activities" or "You might also want to talk to your doctor about…" When patient scenarios deviated from clinical practice guidelines, model responses were less accurate, producing generalized advice that did not account for patients' specific needs. Very general advice came in forms like "It's important to understand how to take care of your back," "If you feel pain, take a break. Resting when it hurts can help prevent more pain later," or "It's essential to stay active."

Another critical finding was the models were unable to offer detailed clinical advice. While the education materials often suggested physical therapy and lifestyle adjustments, these recommendations need more specificity. For example, the responses did not include detailed information on the type, frequency, or intensity of exercises, even though they were rated highly in completeness. Additionally, the suggestions for physical therapy were based on whether the patient felt they would be helpful, rather than on established clinical protocols. Responses looked like "If you're not sure about your lifting technique, ask a physical therapist for help." Also, due to our few-shotting approach, responses predominantly suggested lifting mechanics and desk ergonomics, neglecting other aspects of care like manual therapy. These shortcomings make it difficult to differentiate between the models tested, detracting from the overall completeness of the responses. Even those utilizing RAG approaches did not meet the expectations for clinical application.

Furthermore, redundancy issues were noted. Models repeated patient history unnecessarily, which added little value and detracted from the clarity and usefulness of responses. The responses that were rated more highly provided brief explanations of low back pain, were written in complete sentences, and described exercises in relatively greater detail. Overall, the GenAI models demonstrate potential for providing general information about low back pain, but clinical utility remains limited. The model is more suitable for general public-facing resources such as posted flyers rather than for clinical use, as it struggles to offer detailed, individualized care recommendations.

**Limitations**

Several limitations of this study were identified. First, we used synthetic patient data instead of real patient data. Though synthetic data can mimic real-world patterns to an extent, it lacks the complexity and variability found in actual patterns. This limits the authenticity and representativeness of the results to the real world. Second, the knowledge base used during retrieval was created with input from a single physical therapist (author AB). While the physical therapist's expertise was valuable, the reliance on a single source of clinical knowledge may result in a knowledge base that does not cover the wide spectrum of patient needs. Expanding input to include multiple clinicians would increase the comprehensiveness and reliability of the knowledge base. Lastly, the evaluation process itself presented challenges. The categorical rankings did not always align with the clinical judgments. Minor issues could disproportionately lower a response's overall score. The ambiguity in definitions of certain criteria, such as "completeness," may have biased ratings toward lower scores. Also, a low ICC between the evaluators for both accuracy and completeness limits the reliability of the evaluation process. Notwithstanding these limitations, this study has shown that RAG improves redundancy, accuracy, completeness, and readability over base LLMs. Future studies should consider using real patient data, create a more comprehensive knowledge base, and aim to generate more detailed content.

**Conclusion**

The application of RAG-based GenAI has yielded promising results, particularly when integrated with few-shot learning techniques, for providing personalized patient education materials. Our approach demonstrated superior performance compared to base LLMs in terms of accuracy, completeness, and readability, with reduced redundancy, suggesting a promising tool for enhancing patient education. Despite these encouraging outcomes, our analysis also highlights that the materials generated by GenAI are not yet suitable for clinical implementation. This study emphasizes the potential of AI-driven models utilizing RAG to improve patient education for individuals with LBP while recognizing the significant challenges in ensuring clinical relevance and content specificity.

**Acknowledgment**

The authors would like to acknowledge support from the University of Pittsburgh, Clinical and Translational Science Institute, the School of Health and Rehabilitation Sciences Dean's Research and Development Award, and the National Institutes of Health through Grants UL1TR001857, U24TR004111, and R01LM014306.